\documentclass[lettersize,journal]{IEEEtran}
\usepackage{amsmath,amsfonts}
\usepackage{algorithmic}
\usepackage{algorithm}
\usepackage{array}
\usepackage[caption=false,font=normalsize,labelfont=sf,textfont=sf]{subfig}
\usepackage{textcomp}
\usepackage{stfloats}
\usepackage{url}
\usepackage{verbatim}
\usepackage{graphicx}
\usepackage{cite}
\usepackage{times}
\usepackage{soul}
\usepackage{url}
\usepackage[hidelinks]{hyperref}
\usepackage[utf8]{inputenc}
\usepackage[small]{caption}
\usepackage{graphicx}
\usepackage{amsmath}
\usepackage{amsthm}
\usepackage{booktabs}
\usepackage[switch]{lineno}
\usepackage{multirow} 
\usepackage{amssymb}
\usepackage{bm}
\usepackage[table]{xcolor}
\usepackage{stfloats}
\usepackage{placeins}

\usepackage{wrapfig}

\begin{document}

\title{Contrastive
Augmented Graph2Graph Memory Interaction for Few Shot Continual Learning}

\author{Biqing Qi\textsuperscript{1,2,4}, Junqi Gao\textsuperscript{3}, Xingquan Cheng\textsuperscript{3}, Dong Li\textsuperscript{3}, Jianxing Liu\textsuperscript{1},  ~\IEEEmembership{Senior Member,~IEEE,} Ligang Wu\textsuperscript{1}, ~\IEEEmembership{Fellow,~IEEE,} Bowen Zhou\textsuperscript{1,2,4},~\IEEEmembership{Fellow,~IEEE}

\thanks{
\textsuperscript{1}Department of Control Science and Engineering, Harbin Institute of Technology, Harbin, P. R. China; \textsuperscript{2}Department of Electronic Engineering, Tsinghua University, Beijing, P. R. China; \textsuperscript{3}School of Mathematics, Harbin Institute of Technology, Harbin, P. R. China;
\textsuperscript{4}Frontis.AI, Beijing, P. R. China.
( Emails: qibiqing7@gmail.com; gjunqi97@gmail.com; xinquanchen0117@gmail.com;  arvinlee826@gmail.com; jx.liu@hit.edu.cn; ligangwu@hit.edu.cn; zhoubowen@tsinghua.edu.cn).

Corresponding authors: Bowen Zhou and Ligang Wu.
}

}

\maketitle

\begin{abstract}
Few-Shot Class-Incremental Learning (FSCIL) has gained considerable attention in recent years for its pivotal role in addressing continuously arriving classes. 
However, it encounters additional challenges. The scarcity of samples in new sessions intensifies overfitting, causing incompatibility between the output features of new and old classes, thereby escalating catastrophic forgetting. A prevalent strategy involves mitigating catastrophic forgetting through the Explicit Memory (EM), which comprise of class prototypes. However, current EM-based methods retrieves memory globally by performing Vector-to-Vector (V2V) interaction between features corresponding to the input and prototypes stored in EM, neglecting the geometric structure of local features. This hinders the accurate modeling of their positional relationships. To incorporate information of local geometric structure, we extend the V2V interaction to Graph-to-Graph (G2G) interaction. For enhancing local structures for better G2G alignment and the prevention of local feature collapse, we propose the Local Graph Preservation (LGP) mechanism. Additionally, to address sample scarcity in classes from new sessions, the Contrast-Augmented G2G (CAG2G) is introduced to promote the aggregation of same class features thus helps few-shot learning. Extensive comparisons on CIFAR100, CUB200, and the challenging ImageNet-R dataset demonstrate the superiority of our method over existing methods.
\end{abstract}


\section{Introduction}
\IEEEPARstart{D}{eep} neural networks (DNNs) have demonstrated exceptional performance in downstream tasks like image classification \cite{HeZRS16} and segmentation \cite{HeGDG17}. However, when faced with data from new classes, existing models struggle to consistently adapt.
In contrast, humans continually acquire new knowledge without forgetting the old. Inspired
by this, Class-Incremental Learning (CIL) is introduced to facilitate deep models' adaptation to new class data without erasing prior knowledge \cite{kirkpatrick2017overcoming,SahaG021,pham2021dualnet,mallya2018piggyback}.
However, CIL assumes each new class has ample training data, which is often impractical due to real-world data collection costs.
Consequently, we focus on a more practical setting, few-shot class-incremental learning (FSCIL), designed to learn new classes with a limited number of samples \cite{tao2020few,cheraghian2021semantic}.

\begin{figure}[t]

\centering  
\includegraphics[width=0.5\textwidth]{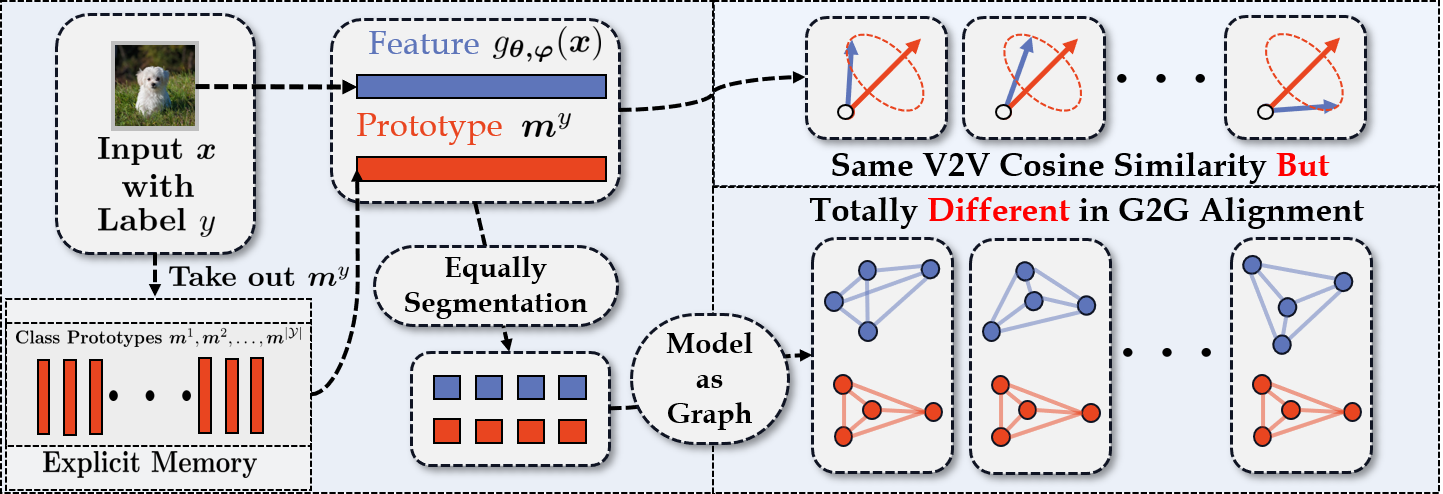}
\caption{Diagram illustrating our motivation. In metric-dependent vector-to-vector alignment, there can be multiple local alignment relationships between the features used for interaction and prototypes in explicit memory, resulting in the same Euclidean distance/cosine similarity between vectors. This leads to imprecise modeling of the positional relationship between features and class prototypes of global vector-to-vector interaction. However, G2G alignment introduces stronger structural constraints, enabling more accurate modeling of the positional relationship between features and class prototypes.}
\label{Fig1}

\end{figure}
However, FSCIL still faces new challenges. Limited availability of samples from new classes in the new session leads to the overfitting on new classes, while restricting the model's generalization on the new session's classes, at the same time, in the case of extremely few samples, the need to effectively learn new classes further exacerbates the difficulty of maintaining knowledge about old classes \cite{tao2020few}, which aggravates catastrophic forgetting \cite{hou2019learning}.

To better address the issues mentioned above, many types of methods have been developed, including incorporate dynamic network structures during training, utilizing base data to help the model adapt in advance to incremental sessions, 
or leveraging a pre-trained feature extractor with Explicit Memory (EM) set by class prototypes and utilize the features for interaction (interactive features) to perform retrieval from EM. The EM-based methods has shown significant promise \cite{HerscheKCBSR22} because it allows more flexible addition of new classes without retraining the classification heads \cite{GravesWRHDGCGRA16,KleykoKRSR23}. More importantly, it effectively alleviate the problem of incompatibility between new and old class features that causes new class features to overwrite old feature classes, because it allows to explicitly maintain the prototypes learned from old sessions. By rehearsing a portion of samples or representations from old sessions, it's more easier to align features from old classes with their related prototyes. This aids in recovering the inter-class feature structure of old classes and mitigating catastrophic forgetting.

However, current EM-based FSCIL methods \cite{santoro2016meta,zhou2022forward,song2023learning,yang2023neural} training only utilizes global feature information, relies solely on the global Vector-to-Vector (V2V) interaction between the entire feature vector corresponding to the input image and prototypes stored in EM for memory retrieval. These methods \cite{santoro2016meta,zhou2022forward,song2023learning,yang2023neural} overlook the distance relationships like euclidean distance and cosine similarity between local features. 

 Based on the aforementioned motivation, we extend the feature-prototype interaction for memory retrieval to a Graph-to-Graph (G2G) interaction in this work. We discretize the extracted features and corresponding prototypes into local features and local prototypes, construct separate graphs based on the euclidean distance between local features and local prototypes, and employ a graph attention mechanism for feature-prototype alignment. This enables us to use node features and positional relationships in the graph to constrain node-level alignment between features and prototypes, enforcing edge-level alignment. Consequently, local geometric structure are introduced into the modeling of feature-prototype interaction, resulting in more accurate retrieval. With stronger geometric constraints, G2G interaction ensures more accurate restoration of features of old classes during rehearsal, thereby mitigating catastrophic forgetting. Notably, our method achieves state-of-the-art (SOTA ) performance by replaying only one image per class, reducing storage overhead.

To further constrain the local positional relationship for more stable local geometric structure \cite{qi2023improving}, we introduce the Local Graph Preservation (LGP) mechanism for G2G. LGP decouples local features to obtain more rich representations, preventing different local features from collapsing into a single representation. More importantly, it introduces more constraint on the local structure of features, enabling more stable interaction modeling and more accurate retrieval. 
Furthermore, to improve the few-shot generalization ability of G2G alignment strategy, the Contrastive Augmented Graph2Graph (CAG2G) interaction is introduced, which models the pre- and post-augmented image features in the same graph and performs graph-to-graph alignment to promote the aggregation of same class features \cite{Chen2020ASF,Huang2021TowardsTG} thus benefits the few-shot learning ability of G2G. 
Recently, the benefits of pre-trained models in low-resource generalization have gained prominence.
However, there has been insufficient exploration in FSCIL setting. 
Therefore, in this paper, we first choose the VIT model to explore its potential applications in FSCIL. 

We conduct comprehensive experiments on CIFAR-100, CUB200, and the challenging ImageNet-R dataset. The experimental results demonstrate that our method outperforms various competitive baselines, confirming the effectiveness of our method.

In summary, our contributions are as follows:
\begin{itemize}
    \item We propose the \textbf{G2G} interaction for memory retrieval, which constructs a graph-to-graph interaction that intorduces euclidean distance between local features and local prototypes, enabling more precise positional relationship modeling therefore easier to recover old interactive class features and reduces catastrophic forgetting.
    \item We design the \textbf{LGP} mechanism to further constrain local positional relationships between features and prototypes thus enhance local structures, which enables more stable interaction modeling and more accurate retrieval.
    \item We propose the effective \textbf{CAG2G}, which further enhances the few-shot generalization ability of graph-to-graph alignment by incorporating contrastive information in aligning the pre- and post-augmented image features in G2G to promote the aggregation of same interactive class features.
\end{itemize}

\section{Related Work}
\textbf{Class Incremental Learning.} Class Incremental Learning (CIL) is a more practical learning method that requires the model to retain previously learned knowledge while learning new classes. CIL methods can be mainly divided into three types. The first type adds regularization \cite{kirkpatrick2017overcoming,zenke2017continual,hinton2015distilling,li2017learning,yu2020semantic, 10083158} so that model parameters that contribute to previous knowledge didn't change significantly. The second type argues that when acquiring new skills, the replaying of samples from previous session \cite{hou2019learning,castro2018end,rebuffi2017icarl, 9939005, hu2022curiosity, liu2020fast} or produced by generative models \cite{cong2020gan} could effectively mitigate the issue of forgetting. The third type forms a dynamic model architecture by constructing session-specific parameters \cite{liu2021adaptive,qin2021bns,qi2024interactive} to adapt to new classes. However, in real world applications, due to the limitation of data cost, only few samples were usually available for new classes, which limits the application of CIL.

\noindent \textbf{Few-shot Class Incremental Learning.} Few-shot Class Incremental Learning (FSCIL) \cite{tao2020few,zhang2021few}, differs from  CIL in that it necessitates training the model on a limited number of samples while learning on new sessions. In other words, the model must acquire new knowledge from a small set of new samples without forgetting previously learned knowledge.
Influenced by the CIL method, a prevalent method incorporated a dynamic network structure, which allowed the model to adapt its network architecture based on the input data characteristics during runtime. TOPIC \cite{tao2020few}, employed neural gas network for preserving the topology in the embedding space, thereby preventing the forgetting of prior knowledge. CEC \cite{zhang2021few}, which model the learned parameters on previeus sessions by graph attention model to facilitate the adaptation of previous model parameters to new sessions. LEC-Net \cite{yang2021learnable} expanded and compressed network nodes to mitigate feature drift and network overfitting. Additionally, distillation techniques were leveraged in \cite{dong2021few,cheraghian2021semantic,zhao2023few} to fine-tuned new session data and reduced the risk of forgetting.

Another popular methods focus on leveraging th observed data to help models adapt to incremental sessions ahead of time. SPPR \cite{zhu2021self} enhanced the scalability of feature representation by forcing features to adapt to various stochastic simulation incremental processes. F2M \cite{shi2021overcoming} found flat local minima in the basic training process, and fine-tuned model parameters in the flat area to adapt to new sessions. LIMIT \cite{zhou2022few} synthesized fake FSCIL sessions from the base data set and builds a generalizable feature space for unseen classes through meta-learning.
Recent methods using EM have gradually became competitive due to their smaller computational cost and better inter-class discrimination. They save prototypes of old classes or create virtual prototypes to explicitly control the intra-class discrimination between old and new classes. DBONet \cite{guo2023decision} combined virtual classes with Prototype vectors for other classes separately. FACT \cite{zhou2022forward} and SAVC \cite{song2023learning} used virtual prototypes to occupy the embedding space, thus reserving space in advance for new classes in the future. Meta-MANN \cite{santoro2016meta} separated the memory storage part from the information processing part, and designed a meta-learner to learn memory storage and retrieval to achieve adaptive interaction between network and explicit memory. GKVM \cite{kleyko2022generalized} introduced generalized key-value memory, and retrieves the memory by querying the key to obtain the corresponding value. C-FSCIL \cite{HerscheKCBSR22} was designed with rewritable, dynamically growing memory and formed a content-based attention mechanism through cosine similarity for query and access in EM. However, current EM-based methods only model the alignment between features and class prototypes globally based on some distance measure, neglecting the geometric structure between local features and local prototypes which hinders accurate modelling of the positional relationship between features and prototypes.

\vspace{-5pt}
\section{Preliminary}
\subsection{Problem Set-up}
In the FSCIL task, training sets (referred to as sessions) $\mathcal D^{(1)}, \mathcal D^{(2)}, \dots, \mathcal D^{(T)}$ are sequentially provided. Each session $\mathcal D^{(t)}:= \{(\boldsymbol x_i^{(t)},y_i^{(t)})\}_{i=1}^{|\mathcal{D}^{(t)}|}$, $1\le t\le T$, consists of input images $\boldsymbol x\in\mathcal X$ and corresponding ground-truth labels $y\in \mathcal Y$. Here, $\mathcal Y^{(t)}$ represents the set of labels for classes (referred to as ways) in the $t$-th session, satisfying $\mathcal Y^{(t)}\cap \mathcal Y^{(s)} = \emptyset$ for $\forall s\ne t$. The first session, $\mathcal Y^{(1)}$, is termed the base session, encompassing more classes than subsequent sessions and providing ample samples for each class in $\mathcal Y^{(1)}$ to facilitate effective representation learning for improved classification. Subsequent sessions $\mathcal Y^{(2)}, \dots, \mathcal Y^{(T)}$ offer $c$ classes (where $c$ is much smaller than $|\mathcal Y^{(1)}|$) and $k$ samples per class for training, commonly referred to as $c$-way $k$-shot. For sessions $t$ where $2 \le t \le T$, only $\mathcal D^{(t)}$ is accessible, and the training sets of previous sessions $\mathcal{D}^{(1)}, \dots, \mathcal{D}^{(t-1)}$ are unavailable. The rehearsal-based strategy involves storing a small memory buffer comprising samples from previous sessions for rehearsal. This buffer is added during subsequent session training to aid the model in recalling knowledge from previous sessions. The test set $\mathcal W^{(t)}$ for each session comprises samples from the classes included in the current session. After training on session $t$, the model undergoes evaluation on the evaluation set $\mathcal{E}^{(t)}=\cup_{k=1}^{t} \mathcal W^{(k)}$, which unifies the test sets from the current session and all preceding sessions.
\subsection{FSCIL with Explicit Memory}
EM strategy for FSCIL \cite{HerscheKCBSR22,song2023learning} usually establish an incremental EM consists of class prototypes, denoted as $\mathcal M={\boldsymbol m^1, \boldsymbol m^2, \dots}$. This incremental EM automatically assigns a trainable class prototype $\boldsymbol m^y \in \mathbb R^{d_m}$ where $d_m$ represents the dimension of $\boldsymbol{m}$ for each newly introduced class $y$, adding it to $\mathcal M$. The set of seen classes is denoted as $\mathcal Y_{\mathcal M}$. During the inference stage, the model retrieves prototypes from $\mathcal M$ using the output features, ensuring retrieval is limited to only seen classes. This is in contrast to generating predictions for unseen classes, as fixed classification head might do. To implement an interaction mechanism for retrieval, a trainable interacter $g_{\boldsymbol\theta}(\cdot)$ parameterized by $\boldsymbol\theta$ is introduced. For a given input $\boldsymbol x$, first perform feature extraction: $\boldsymbol h = f_{\boldsymbol\varphi}(\boldsymbol x)\in\mathbb R^{d_h \times L}$ using a frozen pre-trained feature extractor, which remain fixed during incremental sessions. Subsequently, $\boldsymbol h$ is fed into the interacter $g_{\boldsymbol\theta}(\cdot)$ to obtain the interactive feature $\boldsymbol\xi = g_{\boldsymbol\theta}(\boldsymbol{h})$. Finally, a retrieval operation $\mathcal R_{\mathcal M}(\boldsymbol\xi)$ is performed to obtain the memory retrieval result $\hat y$. For simplicity, we denote $g_{\boldsymbol\theta, \boldsymbol\varphi}=g_{\boldsymbol\theta}\circ f_{\boldsymbol\varphi}(\cdot)$, where the symbol $\circ$ represents the composition operation. Typically, retrieval is based on a distance metric $d(\cdot, \cdot)$, i.e.:

\begin{equation}
    \mathcal R_{\mathcal M}(\boldsymbol x) = \arg\min_{y\in \mathcal Y_{\mathcal M}} d(g_{\boldsymbol\theta,\boldsymbol \varphi}(\boldsymbol x),\boldsymbol m^y),
\end{equation}
and $g_{\boldsymbol\theta}$ is usually instantiated using an MLP. However, such retrieval strategy only leverages the global interaction between $g_{\boldsymbol\theta,\boldsymbol \varphi}(\boldsymbol x)$ and $\boldsymbol m$, neglecting the local geometric structure information of $g_{\boldsymbol\theta,\boldsymbol \varphi}(\boldsymbol x)$ and $\boldsymbol m$ individually. Consequently, $g_{\boldsymbol\theta,\boldsymbol \varphi}(\boldsymbol {x_1})$ and $g_{\boldsymbol\theta,\boldsymbol \varphi}(\boldsymbol {x_2})$ are considered equivalent during retrieval on $\boldsymbol m$, even if $\boldsymbol {x_1} \ne \boldsymbol {x_2}$, due to $d(g_{\boldsymbol\theta,\boldsymbol \varphi}(\boldsymbol {x_1}), \boldsymbol m) = d(g_{\boldsymbol\theta,\boldsymbol \varphi}(\boldsymbol {x_2}), \boldsymbol m)$. This limitation hinders an accurate representation of the positional relationship between them through this interaction modeling mechanism.

\vspace{-5pt}
\section{Methodology}
\label{Methodology}
\subsection{Graph2Graph Memory Interaction}

\begin{figure*}[htbp]

\centering 
\includegraphics[width=0.9\textwidth]{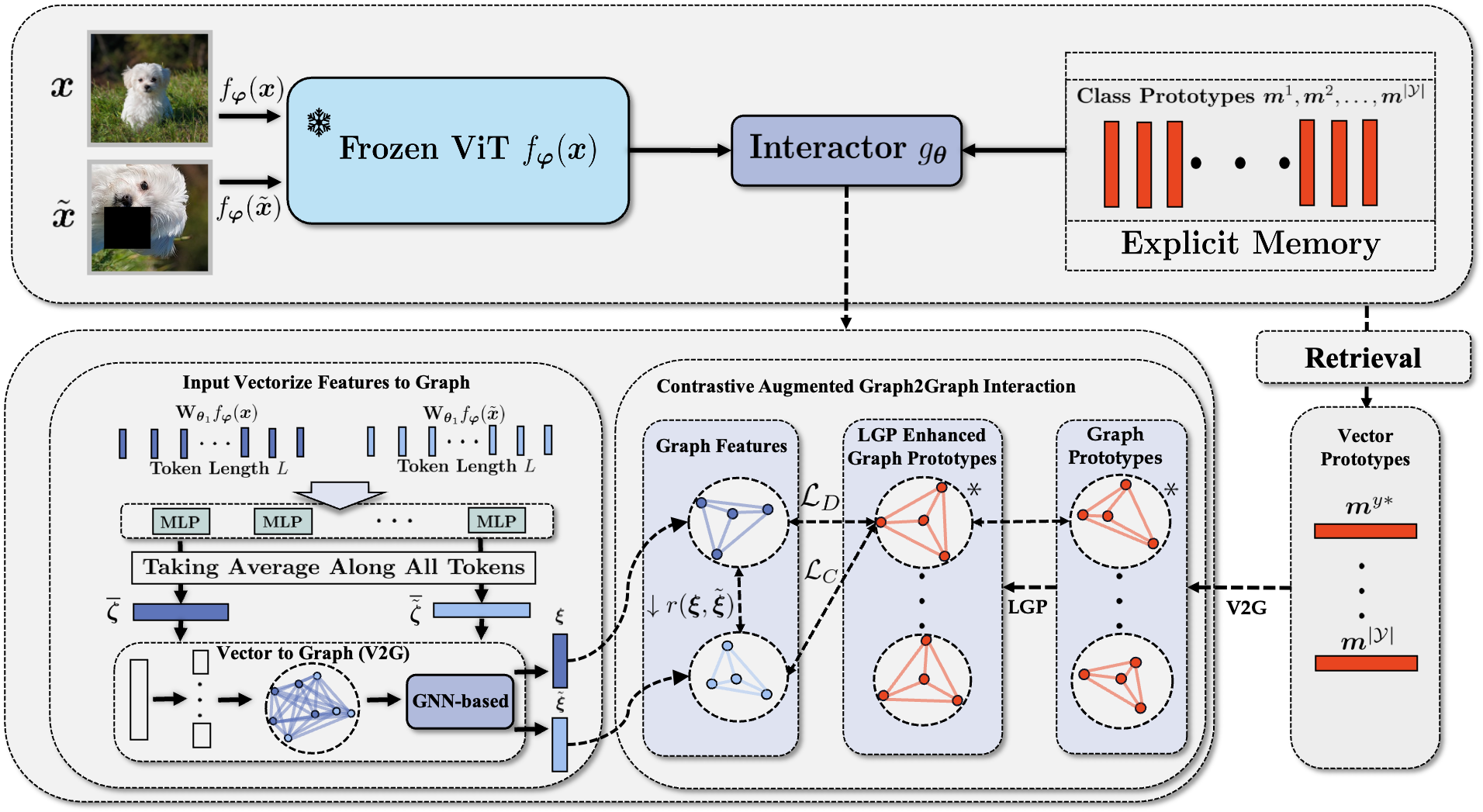}
\caption{
Overview of our method: For each input image $\boldsymbol{x}$, we feed it, along with its enhanced view $\tilde{\boldsymbol{x}}$, into the pre-trained ViT. The resulting features $f_{\boldsymbol{\varphi}}(\boldsymbol{x})$ and $f_{\boldsymbol{\varphi}}(\tilde{\boldsymbol{x}})$ are then input into the interactor $g_{\boldsymbol{\theta}}$ for G2G interaction. During the training phase, we conduct G2G alignment between the features of both the original and augmented views output from $g_{\boldsymbol{\theta}}$. This encourages improved intra-class concentration, simultaneously aligning the output features of the two views with the ground-truth label (denoted as $y^*$) through G2G.}

\label{Fig2}

\end{figure*}

\noindent In order to introduce local geometric structure into the interaction, we first consider the output features locally, as shown in Fig. \ref{Fig2}. Denoting $\boldsymbol h = [\boldsymbol {h}_1^\top, \boldsymbol{h}_2^\top, \dots, \boldsymbol {h}_S^\top]^\top$, where $\boldsymbol {h}_s\in\mathbb R^{\frac{d_h}{S}\times L}$ represents the $s$-th local feature after equally partitioning. Here we choose pre-trained ViT as the feature extractor $f_{\boldsymbol{\varphi}}(\cdot):\mathcal X\rightarrow \mathbb R^{d_h\times L}$ to explore the performance of ViT architecture in FSCIL tasks, where $L$ is the token length of the ViT output. We set a learnable matrix $\mathbf W_{\boldsymbol {\theta}_1}\in\mathbb R^{d_h\times d_h}$, parameterized by $\boldsymbol {\theta}_1$, to adjust the output features, denote as $\bm {\hbar} = \mathbb L(\boldsymbol h) := \mathbf W_{\boldsymbol {\theta}_1} \boldsymbol h$. Then we consider the local information of $\bm \hbar$. By setting a learnable MLP module $\mathcal T_{\bm\theta_{2, s}}$ for each local feature $\bm\hbar_s$, which parameterized by $\bm\theta_{2, s}$, we continue to perform transformations on $\bm\hbar$ to obtain $\bm\zeta = [\mathcal T_{\bm\theta_{2, 1}}(\bm\hbar_1)^\top, \mathcal T_{\bm\theta_{2, 2}}(\bm\hbar_2)^\top, \dots, \mathcal T_{\bm\theta_{2, S}}(\bm\hbar_S)^\top]^\top$. Here, we also set an equally splited hidden dimension for each $\mathcal T_{\bm\theta_{2, s}}$, so the total number of parameters $\sum_{s=1}^S|\bm\theta_{2, s}|$ decreases as the number of partitions $S$ increases. This does not bring additional parameter storage burden, but instead reduces the complexity of the model. 

Next, we average $\bm\zeta$ across the token dimension to obtain $\bm{\bar\zeta} = [\bm{\bar\zeta}_1^\top, \bm{\bar\zeta}_2^\top, \dots, \bm{\bar\zeta}_S^\top]^\top\in\mathbb R^{d_\zeta}$, where $\bm{\bar\zeta}_s\in\mathbb R^{\frac{d_\zeta}{S}}$. Using $\bm{\bar\zeta}$, we extend the original vector-to-vector interactions to graph-to-graph interactions that incorporate geometric structure between local features. Specifically, we model different local features $\bm {\bar\zeta}_i$ into a weighted graph with an local adjacency matrix $\mathbf A$. The elements of $\mathbf A$ satisfy $\mathbf A_{i,j} = 1/ \exp(\|\bm{\bar\zeta}_i-\bm{\bar\zeta}_j\|_2)$, reflecting the geometric distances between different local features.

\subsection{G2G-GNN for Memory Interaction}
By utilizing the constructed local feature weighted graph, we adopt several representative graph neural network models $\mathcal A(\cdot):\mathbb R^{d_\zeta}\rightarrow \mathbb R^{d_\xi}$ to capture the geometric structure of the output features.

\paragraph{G2G-GCN} Graph Convolutional Neural Networks (GCN) \cite{kipf2016semi} is a computationally simple and efficient method that has shown significant advantages in the field of graph modeling. By utilizing a two-layer graph convolutional algorithm, GCN can capture features that contain local interactive geometric information. Compared with vanilla GCN, the G2G-GCN constructs a unique graph for each sample in the batch. As a result, the input to G2G-GCN contains structural information and node features from multiple graphs. For each graph in the G2G-GCN model, a two-layer graph convolution operation is applied to capture the local geometric structure of the output features:
\begin{equation}
    \mathcal{A}(\bm\bar{\zeta}) = {\rm{softmax}}(\mathbf{A}{\rm{ReLU}}(\mathbf{A}\bm\bar{\zeta}{\mathbf{W}^{(0)})\mathbf{W}^{(1)}}),
\end{equation}

where ReLU denotes the ReLU activition function, $\mathbf{W}^{(i)}\in \mathbb{R}^{d_{\mathcal{S}}\times h}$ is learnable weight matrix in $i$-th layer and $h$ is output dimension.

\paragraph{G2G-GAT} Graph Attention Network (GAT) \cite{Velickovic2017GraphAN} integrates attention mechanisms into the Graph Neural Network (GNN) framework, thereby facilitating advanced fusion of feature correlations among nodes in the model. Similar to G2G-GCN, G2G-GAT possesses the capability to concurrently handle multiple graphs. In the context of employing the G2G-GAT module for inter-graph interactions, two steps procedure predominates. The initial step involves the specification of a set of learnable parameterized vectors $\bm a \in\mathbb R^{2d_\zeta/S}$ and matrices $\bm W_\zeta\in\mathbb R^{\frac{d_\zeta}{S}\times \frac{d_\zeta}{S}}$ for the computation of similarity coefficients between local features:
\begin{equation}
    e_{i,j} = \bm a^\top[\mathbf W_\zeta\bm \bar{\zeta}_{i} \|\mathbf W_\zeta\bm \bar{\zeta}_{j}],
\end{equation}
where $\|$ denotes the concatenation operation. Then, for each graph, we utilize the similarity coefficients $e_{i,j}$ and the weighted adjacency matrix $\mathbf A$ to construct attention scores:
\begin{equation}
    \alpha_{i,j} = \mathbf A_{i,j}*\frac{\exp\left(\text{LeakyReLU}(e_{i,j})\right)}{\sum_{j\in S}\exp\left(\text{LeakyReLU}(e_{i,j})\right))}.
\end{equation}
Next, the local feature aggregation operation is performed by attention scores to obtain the output of $\mathcal A(\cdot)$:
\begin{equation}
    \mathcal A(\bm\bar\zeta) = \sigma\left( \left[\sum_{j=1}^S\alpha_{1,j}\bm\bar{\zeta}_{j}^\top\mathbf W_\zeta^\top, \dots, \sum_{j=1}^S\alpha_{S,j}\bm\bar{\zeta}_{j}^\top\mathbf W_\zeta^\top\right]^\top \right),
\end{equation}
where $\sigma(\cdot)$ represents a pre-defined activition function. 
The next step is to set $K$ independent $\mathcal A^i$ to construct multi-head attention and obtain the final features used for interactions, denote as $\bm\xi = \mathcal A^{o}\left(\|_{i=1}^{K}{\mathcal A^i\left(\bm\zeta\right)}\right)$, where $\mathcal A^o$ is used for executing feature aggregation on the concatenated features.

\paragraph{G2G-PairNorm} To avoid limiting the expressive ability of the model due to the feature distances between nodes being too close during training, Zhao\cite{zhao2019pairnorm} proposed the PairNorm method. This method achieves this goal by adding regularization to the output of each graph convolution layer. Let $\bm\bar{\zeta}$ be the original node input of PairNorm, and $\dot{\bm\bar{\zeta}}$ be the output of the graph convolution, which is calculated as shown in Equation (2). The essence of PairNorm is to ensure that the total pairwise squared distance (TPSD) of the input $\bm\bar{\zeta}_{i}$ in the graph is equal to the TPSD of $\dot{\bm\bar{\zeta}}_{i}$, where TPSD is defined as follows:
\begin{equation}
    TPSD(\dot{\bm\bar{\zeta}}) = \sum_{i,j=0,...,\mathcal{S}}\parallel \dot{\bm\bar{\zeta}}_{i}- \dot{\bm\bar{\zeta}}_{j} \parallel^{2}_{2}.
\end{equation}
The aim of the G2G-PairNorm method is to ensure that the TPSD of $\dot{\bm\bar{\zeta}}$ is equal to that of $\bm\bar{\zeta}$, namely:

\begin{equation}
\begin{aligned}
& \sum_{(i,j)\in{\mathcal{E}}}\parallel \dot{\bm\bar{\zeta}}_{i} - \dot{\bm\bar{\zeta}}_{j} \parallel^{2}_{2} + \sum_{(i,j)\notin{\mathcal{E}}}\parallel \dot{\bm\bar{\zeta}}_{i} - \dot{\bm\bar{\zeta}}_{i} \parallel^{2}_{2} = \\
& \sum_{(i,j)\in{\mathcal{E}}}\parallel {\bm\bar{\zeta}}_{i} - {\bm\bar{\zeta}}_{j} \parallel^{2}_{2} + \sum_{(i,j)\notin{\mathcal{E}}}\parallel {\bm\bar{\zeta}}_{i} - {\bm\bar{\zeta}}_{i} \parallel^{2}_{2}.
\end{aligned}
\end{equation}

Here, $\mathcal{E}$ denotes the set of edges in $i$-th input graph. In G2G module, PairNorm can handle multiple graph data simultaneously. 
\paragraph{G2G-GCNII}
GCN stacks $k$ layers of graph convolutions to simulate a $K$th-order polynomial filter with fixed coefficients $(\sum^{K}_{l=0}\theta_{l}\mathbf{L}^{l})\bm\bar{\zeta})$ on the graph domain $\mathcal{G}$ \cite{chen2020simple}. Here, $L$ is the Laplacian matrix. Due to the inherent properties of GCN, the expressive ability of multi-layer GCN models is limited by the fixed coefficients. To address this limitation and enable GCN to express $K$th-order polynomial filters with arbitrary coefficients, Ming \cite{chen2020simple} proposed the GCNII method. This method utilizes initial residual connections and identity mapping techniques to solve the aforementioned problem:
\begin{equation}
    \mathbf{H}^{(i+1)} = \sigma(\underbrace{((1-\gamma^{i})\mathbf{A}\mathbf{H}^{(i)}+\gamma^{i}\bm\bar{\zeta})}_{\text{residual connections}}((1-\eta^{i})\mathbf{I}+\eta^{i}\mathbf{W}^{(i)})),
\end{equation}
where $\gamma^{i}, \eta^{i} \in \mathbb{R}$ are hyperparameters, and $\mathbf{W}^{(i)}\in \mathbb{R}^{\mathcal{S}\times \mathcal{S}}$ is the learnable weight matrix for layer $i$.
In G2G-GCNII, we employ 16 layers of GCNII convolution and capture the geometric structure of the output features through the aforementioned process:
\begin{equation}
    \mathcal{A}(\bm\bar{\zeta}) = \mathbf{H}^{16}(\mathbf{H}^{15}(\cdots \mathbf{H}^{2}(\mathbf{H}^{1}(\bm\bar{\zeta})))).
\end{equation}

\paragraph{G2G-GGCN}
To further enhance the model's ability to capture local topological information, Yan \cite{yan2022two} designed the GGCN method. Similar to the improved GNN model in the previously introduced G2G method, it is necessary to enhance the original GGCN, which could only handle a single graph, to a model that can simultaneously handle multiple graphs. In GGCN, a key step is to compute the cosine similarity between nodes in the same graph.
After obtaining the cosine similarity between nodes in each graph, we compute the similarity matrix obtained and further divide it into matrices $\mathbf{S}_{pos}$ and $\mathbf{S}_{neg}$ based on the positive and negative cosine similarities. In addition, for the G2G-GGCN module, we apply two layers of graph convolution operations to each input graph, with the specific form of graph convolution as follows:

\begin{equation}
\begin{split}
    \mathbf{H}^{(l+1)} = & \sigma(\kappa^{l}(\mu_{0}^{l}\mathbf{H}^{(l)}+\mu_{1}^{l}(\mathbf{S}^{l}_{pos}\odot \mathbf{A})\mathbf{H}^{(l)} \\
    & +\mu_{2}^{l}(\mathbf{S}^{l}_{neg}\odot \mathbf{A})\mathbf{H}^{(l)})),
\end{split}
\end{equation}

where $\sigma$ is the nonlinear function Elu. $\kappa^{l}, \mu_{0}^{l}, \mu_{1}^{l}, \mu_{2}^{l} \in \mathbb{R}$ are learnable parameters in $l$-th layer and $\odot$ is element-wise multiplication. In G2G-GGCN, we use 6 layers of graph convolution to capture local geometric relations:
\begin{equation}
    \mathcal{A}(\bm\bar{\zeta}) = \mathbf{H}^{6}(\mathbf{H}^{5}(\cdots \mathbf{H}^{2}(\mathbf{H}^{1}.(\bm\bar{\zeta})))).
\end{equation}

\paragraph{G2G-GraphSage}
GraphSage \cite{hamilton2017inductive} is a graph neural network method that effectively learns node representations by sampling and aggregating features of neighboring nodes, thereby enabling graph node classification and representation learning. For each graph, when computing the embeddings of $k$-th layer of neighboring nodes, we first use the embeddings of the $(k-1)$-th layer of its neighboring nodes to calculate the aggregated representation of $k$-th layer of neighboring nodes $\bm\bar{\zeta}^{k}_{N(i)}$. Then, we connect $\bm\bar{\zeta}^{k}_{N(i)}$ with the $(k-1)$-th layer representation of node $i$ to obtain the representation of node $i$ at layer $k$:
\begin{align}
    &\bm\bar{\zeta}^{k}_{N(i)} = \text{Aggregate}({\bm\bar{\zeta}^{k-1}_{i}, \forall j \in N(i)}) \\
    &\bm\bar{\zeta}_{i}^{k} = \sigma(\mathbf{W}^{k} \cdot \text{Concat}(\bm\bar{\zeta}_{i}^{k-1}, \bm\zeta^{k}_{N(i)})),
\end{align}
where $N(i)$ denotes the neighbors of $i$, $\mathbf{W}^{k}$ is a learnable weight matrix in $k$-th layer. In addition, we adopt the following aggregation method to aggregate information from neighbors:
\begin{equation}
    \bm\bar{\zeta}^{k}_{N(i)} = \text{Mean}(\sigma(\mathbf{W}\bm\bar{\zeta}^{k-1}_{j}+b), \forall j \in N(i)),
\end{equation}
where $\mathbf{W}$ denotes a learnable weight matrix. Therefore, we can obtain the output features of G2G-GraphSage:
\begin{equation}
    \mathcal{A}(\bm\bar{\zeta}) = [\bm\bar{\zeta}_{1}^{2\top},...,\bm\bar{\zeta}_{\mathcal{S}}^{2\top}]^{\top}.
\end{equation}

\paragraph{G2G-GATv2} Brody \cite{brody2021attentive} argued that GAT computes static attention, meaning that for a fixed set of keys, if different queries attend to this set of keys, the ordering of the attention coefficients remains unchanged. This limitation restricts the performance of GAT on graph tasks, and the authors achieved dynamic attention computation through a simple modification to GAT:
\begin{equation}
    e_{i,j} = \bm a^\top\text{LeakyReLU}(\mathbf W_\zeta[\bm\bar{\zeta}_i \|\bm\bar{\zeta}_j]).
\end{equation}
where $\bm a^\top$ is the same to equation (3) and $\mathbf W_\zeta \in \mathbb {R}^{\frac{d_\zeta}{S}\times 2\frac{d_\zeta}{S}}$. Similar to the method of calculating attention scores in G2G-GAT, the method for computing attention scores in G2G-GATv2 is as follows:
\begin{equation}
    \alpha_{i,j} = \mathbf A_{i,j}*\frac{\exp\left((e_{i,j})\right)}{\sum_{j\in S}\exp\left((e_{i,j})\right))}.
\end{equation}
According to the above calculation process, the output of the G2G-GATv2 module can be obtained as follows:
\begin{equation}
    \mathcal A(\bm\bar\zeta) = \sigma\left( \left[\sum_{j=1}^S\alpha_{1,j}\bm\bar{\zeta}_{j}^\top\mathbf W_\zeta^\top, \dots, \sum_{j=1}^S\alpha_{S,j}\bm\bar{\zeta}_{j}^\top\mathbf W_\zeta^\top\right]^\top \right),
\end{equation}
After integrating the above seven GNNs into the G2G Interaction module, the function $g_{\boldsymbol \theta}$ can be uniformly formulated as follows:
\begin{equation}
    g_{\boldsymbol \theta}(\cdot):= \mathcal A \circ \text{Ave}\circ{\mathcal T}\circ\mathbb L(\cdot),
\end{equation}
where Ave represents the operation of taking average and $\mathcal A(\cdot)$ represents the specified G2G-GNN. Utilizing the interaction features $\bm\xi=g_{\bm \theta}(\bm h)$, we design the G2G interactions to perform memory retrieval. Similarly, we divide the interaction features $\bm\xi$ and prototype $\bm m$ equally into local interaction features $\bm\xi_s\in\mathbb R^{\frac{d_\xi}{S}}$ and local prototype vectors $\bm m_s\in\mathbb R^{\frac{d_\xi}{S}}$. We also construct local adjacency matrices for them in the same way, denoted as $\mathbf A^{\bm\xi}$ and $\mathbf A^{\bm m}$, respectively. Utilize the notations above, we design the graph-level dissimilarity metric $r(\cdot, \cdot)$ to simultaneously measure the feature difference and the local geometric structure difference between $\bm\xi$ and $\bm m$:
\begin{equation}
    r(\bm \xi, \bm m) = \|\bm \xi-\bm m\|_2 + \|\mathbf A^{\bm \xi}-\mathbf A^{\bm m}\|_F,
\end{equation}
then our G2G memory retrieval can be represented as:
\begin{equation}
    \mathcal R_{\mathcal M}(\boldsymbol x) = \arg\min_{y\in \mathcal Y_{\mathcal M}} r(g_{\boldsymbol\theta,\boldsymbol \varphi}(\boldsymbol x),\boldsymbol m^y).
\end{equation}
To endow the model with stable G2G retrieval capability during the inference stage, it is necessary to encourage tight G2G alignment during the training phase. Therefore, we introduce the batch-wise G2G prototype contrastive loss $\mathcal L_{G}$ during the training stage to promote G2G prototype alignment and ensure that each sample stays far away from prototypes from other classes appearing in the same batch. Given a batch of samples $\mathcal B = \{(\bm x_i, y_i)\}_{i=1}^{|\mathcal B|}$ and denote $\mathcal Y^{\mathcal B}$ as all the classes observed in $\mathcal B$, $\mathcal L_{G}$ is defined as:
\begin{equation}
    \mathcal L_{G}(\mathcal B) = \frac{1}{|\mathcal B|}\sum_{i=1}^{|\mathcal B|}-\log\left[\frac{\exp\left(-r(g_{\boldsymbol\theta,\boldsymbol \varphi}(\boldsymbol x_i),\boldsymbol m^{y_i})\right)}{\sum_{y\in\mathcal Y^{\mathcal B}}\exp\left(-r(g_{\boldsymbol\theta,\boldsymbol \varphi}(\boldsymbol x_i),\boldsymbol m^{y})\right)}\right].
\end{equation}

After training on each session $t$, we freeze the prototypes corresponding to the classes seen in the current session, retaining their learned structure in subsequent sessions. Our G2G alignment enables more precise modeling of the positional relationship between features and class prototypes. It allows accurate recovery of features from previous classes with only a small subset of samples used for rehearsal, thus effectively alleviates catastrophic forgetting.
\subsection{Contrastive Augmented Graph2Graph
Memory Interaction}

\paragraph{
Local
Graph Preservation Mechanism
 }
To further bolster the stability of G2G interaction while promoting their capacity for few-shot generalization, we begin by focusing on the geometric structure of local features. Considering that: 1) a more clearly defined local geometric structure of prototypes leads to more stable G2G alignment, and 2) decoupling of local features yields richer feature representations, which helps prevent local feature collapse that can impede generalization \cite{BardesPL22}. Thus, we introduce a LGP mechanism by design a Local Decoupling Loss $\mathcal L_D$ to enforce additional decoupling constraints on the structure of local prototype vectors:
\begin{equation}
    \mathcal L_D(\mathcal B) = \frac{1}{|\mathcal B|}\sum_{i=1}^{|\mathcal B|}\sum_{s=1}^S\sum_{1\le j\le S\atop j\ne s}  \frac{\left \langle \boldsymbol{m^{y_i}}_s, \boldsymbol{m^{y_i}}_j \right \rangle}{\left\|\boldsymbol{m^{y_i}}_s\right\|_2\left\|\boldsymbol{m^{y_i}}_j\right\|_2}.
\end{equation}
LGP ensures further constraints on the structure of local prototypes, providing more explicit guidance during the alignment phase, resulting in a more stable G2G alignment. At the same time, it further decouples the local features during the alignment process, leading to richer feature representations.

\paragraph{Contrastive Augmented G2G Interaction}
To maximize the utilization of sample information and further enhance the few-shot generalization ability of G2G, we introduce a contrastive enhancement mechanism for G2G to CAG2G Interaction, which modeling the samples before and after augmentation into the same graph for G2G alignment. Specifically, for the pre-augmented and post-augmented samples $\bm x$ and $\tilde{\bm x}$, we first compute their corresponding $\bm\zeta$ and $\tilde{\bm\zeta}$. Then, we concatenate them to obtain $\bm\zeta^{[\bm x, \tilde{\bm x}]}=[{\bm\zeta}\|\tilde{\bm\zeta}]\in\mathbb R^{2d_\zeta}$ and construct a weighted graph of the concatenated local features. Then we apply the weighted graph attention to obtain the interactive feature $\bm\xi^{[\bm x, \tilde{\bm x}]}\in\mathbb R^{2d_\xi}$. Finally, we extract the corresponding interaction features $\bm \xi$ and $\tilde{\bm\xi}$:
\begin{equation}
    \bm \xi = \bm\xi^{[\bm x, \tilde{\bm x}]}[: d_\xi], \quad \tilde{\bm\xi} = \bm\xi^{[\bm x, \tilde{\bm x}]}[d_\xi:],
\end{equation}
we propose a local graph contrastive loss $\mathcal L_C$ to introduce contrastive information of samples during the alignment process:
\begin{equation}
    \mathcal L_C(\mathcal B)=\mathcal L_{G}(\tilde{\mathcal B})+\frac{1}{|\mathcal B|}\sum_{i=1}^{|\mathcal B|}\left[\|\boldsymbol \xi^i-\tilde{\boldsymbol \xi}^i\|_2^2+\|\mathbf A^{\boldsymbol \xi^i}-\mathbf A^{\tilde{\boldsymbol \xi}^i}\|_F^2\right],
\end{equation}
where $\tilde {\mathcal B}$ represents the batch $B$ after data augmentation, $\bm \xi^i$ and $\tilde{\bm \xi}^i$ represent the interaction features corresponding to samples $\bm x_i$ and $\tilde {\bm x}_i$ obtained from eq.(24). CAG2G facilitates the model to learn more robust feature representations to enhance its few-shot generalization ability.

Therefore, our overall loss is as follows: 
\begin{equation}
    \mathcal L(\mathcal B) = \mathcal L_G(\mathcal B) + \lambda\mathcal L_D(\mathcal B) + \eta\mathcal L_C(\mathcal B),
\end{equation}
where $\lambda$ and $\eta$ are hyperparameters. To ensure that the model structure constrained by LGP can be aligned more stably, we train class prototypes for $I$ iterations in each session and then fix them to reduce alignment instability caused by prototype changes during training.
\section{Experiments}

In this section, we conduct extensive evaluation experiments on three datasets to compare our method with various competitive baselines to validate the effectiveness of our method.
Additionally, we conduct ablation experiments to validate the effectiveness of proposed local graph preservation mechanism and contrastive augmented G2G interaction method.
To ensure a fair comparison, we maintain uniformity in the partitioning of training data during evaluation. 

\begin{table*}[!t]
  \centering
  \scriptsize
    \setlength{\tabcolsep}{2.8mm}
\renewcommand{\arraystretch}{1.0}
\resizebox{0.9\linewidth}{!}{
    \begin{tabular}{cccccccccccc}
    \toprule
    \multirow{2}{*}{Method} & \multicolumn{9}{c}{Acc. in each session (\%) ↑}                                      & \multirow{2}{*}{PD $\downarrow$} & \multirow{2}{*}{\shortstack{Average \\ ACC ↑}}\\
\cmidrule{2-10}          & 0     & 1     & 2     & 3     & 4     & 5     & 6     & 7     & 8     &  & \\
    \midrule
    iCaRL \cite{rebuffi2017icarl} & 64.10  & 53.28  & 41.69  & 34.13  & 27.93  & 25.06  & 20.41  & 15.48  & 13.73  & 50.37  & 32.87 \\
    EEIL \cite{castro2018end}  & 64.10  & 53.11  & 43.71  & 35.15  & 28.96  & 24.98  & 21.01  & 17.26  & 15.85  & 48.25  & 33.79  \\
    NCM \cite{hou2019learning}  & 64.10  & 53.05  & 43.96  & 36.97  & 31.61  & 26.73  & 21.23  & 16.78  & 13.54  & 50.56  & 34.22  \\
    D-Cosine \cite{vinyals2016matching}  & 74.55  & 67.43  & 63.63  & 59.55  & 56.11  & 53.80  & 51.68  & 49.67  & 47.68  & 26.87  & 58.23  \\
    TOPIC \cite{tao2020few} & 64.10  & 55.88  & 47.07  & 45.16  & 40.11  & 36.38  & 33.96  & 31.55  & 29.37  & 34.73  & 42.62  \\
    CEC$^{*}$ \cite{zhang2021few} & 72.99 &	68.85 &	65.04 &	61.13 &	58.03 &	55.53 &	53.20 &	51.17 &	49.05 &	23.94 &	59.44  \\
    MetaFSCIL \cite{chi2022metafscil} & 74.50 &	70.10 &	66.84 &	62.77 &	59.48 &	56.52 &	54.36 &	52.56 &	49.97 &	24.53 &	59.44 \\
    FACT \cite{zhou2022forward} & 74.60 &	72.09 &	67.56 &	63.52 &	61.38 &	58.36 &	56.28 &	54.24 &	52.10 &	22.50 &	62.24 \\
    ERDR \cite{liu2022few} & 74.40 &	70.20 &	66.54 &	62.51 &	59.71 &	56.58 &	54.52 &	52.39 &	50.14 &	24.26 &	60.78  \\
    DBONet \cite{guo2023decision} & 77.81 &	73.62 &	71.04 &	66.29 &	63.52 &	61.01 &	58.37 &	56.89 &	55.78 &	\underline{22.03} &	64.93  \\
    BiDist$^{*}$ \cite{zhao2023few} & 73.55 &	69.08 &	64.67 &	60.64 &	57.06 &	53.92 &	51.93 &	49.84 &	47.18 &	26.37 &	58.65 \\
    SAVC$^{*}$ \cite{song2023learning} & 78.85 &	73.26 &	68.97 &	64.53 &	61.28 &	58.07 &	55.87 &	53.92 &	51.87 &	26.98 &	62.96 \\
    NC-FSCIL \cite{yang2023neural} & 82.52 &	76.82 &	73.34 &	69.68 &	66.19 &	62.85 &	\underline{60.96} &	\underline{59.02} &	\underline{56.11} &	26.41 &	67.50 \\
    
    \midrule
    ViT finetune & 83.84 & 	74.77 & 	69.10 & 	60.74 & 	59.25 & 	48.33 & 	47.56 & 	41.90 & 	39.20 & 	44.64 & 	58.30 \\
    L2P$^*$ \cite{0002ZL0SRSPDP22} & 88.47 & 	43.53 & 	28.73 & 	21.45 & 	17.04 & 	14.02 & 	11.79 & 	9.96 & 	8.81 & 	79.66 & 	27.09 \\
    DualPrompt$^*$ \cite{0002ZESZLRSPDP22} & 89.70 & 	55.23 & 	46.66 & 	35.19 & 	29.26 & 	26.86 & 	24.17 & 	18.10 & 	13.69 & 	76.01 & 	37.67 \\
    \midrule
    \rowcolor{blue!10}LGP-CAG2G(GAT)  & 89.97  & \textbf{86.51}  & \underline{83.80}  & 80.24  & \underline{80.76}  & \textbf{79.18}  & \textbf{78.97}  & \underline{76.32}  & 75.15  & 14.82  & 81.21 \\
     \rowcolor{blue!10}LGP-CAG2G(GraphSage) & 90.02 & 	85.83 & 	\textbf{84.26} & 	\underline{80.57} & 	79.69 & 	79.03 & 	78.76 & 	\textbf{77.49} & 	\underline{75.52} & 	\underline{14.50} & 	\underline{81.24} \\
     \rowcolor{blue!10}\textbf{LGP-CAG2G(GATv2)} & \textbf{90.13} & 	\underline{86.02} & 	\underline{83.97} & 	\textbf{80.73} & 	\textbf{80.80} & 	\underline{79.06} & 	\underline{78.84} & 	\underline{77.27} & 	\textbf{76.06} & 	\textbf{14.07} & 	\textbf{81.43} \\
    \rowcolor{blue!10}LGP-CAG2G(GCN) & 89.87 & 	82.77 & 	77.35 & 	72.16 & 	67.06 & 	63.49 & 	58.96 & 	55.82 & 	52.74 & 	37.13 & 	68.91 \\
    \rowcolor{blue!10}LGP-CAG2G(GCNII) & 90.02 & 	82.63 & 	75.06 & 	70.01 & 	65.63 & 	60.74 & 	58.43 & 	54.45 & 	51.63 & 	38.39 & 	67.62 \\
    \rowcolor{blue!10}LGP-CAG2G(PairNorm) & \underline{90.07} & 	81.52 & 	75.94 & 	69.65 & 	65.09 & 	60.61 & 	57.33 & 	54.18 & 	51.95 & 	38.12 & 	67.37 \\
    \rowcolor{blue!10}LGP-CAG2G(GGCN) & 90.05 & 	81.89 & 	75.65 & 	69.77 & 	65.27 & 	60.65 & 	57.70 & 	54.27 & 	51.84 & 	38.21 & 	67.45 \\
    \bottomrule
    \end{tabular}%
    }
    \caption{Comparison with SOTA methods on CIFAR100 dataset for FSCIL. $*$ are reproduced following the settings of the original paper. The remaining experimental results come from CEC\protect\cite{zhang2021few}, NC-FSCIL\protect\cite{yang2023neural} or the results reported in the original paper.}
  \label{tab:CIFAR100}%

\end{table*}%

\begin{table*}[!t]
  \centering
    \setlength{\tabcolsep}{2.8mm}
\renewcommand{\arraystretch}{1.0}
\resizebox{0.9\linewidth}{!}{
    \begin{tabular}{cccccccccccccc}
    \toprule
    \multirow{2}{*}{Method} & \multicolumn{11}{c}{Acc. in each session (\%) ↑}                                      & \multirow{2}{*}{PD $\downarrow$} & \multirow{2}{*}{\shortstack{Average \\ ACC ↑}}\\
\cmidrule{2-12}          & 0     & 1     & 2     & 3     & 4     & 5     & 6     & 7     & 8    & 9 & 10 &  & \\
    \midrule
    iCaRL \cite{rebuffi2017icarl} & 68.68 &	52.65 &	48.61 &	44.16 &	36.62 &	29.52 &	27.83 &	26.26 &	24.01 &	23.89 &	21.16 &	47.52 &	36.67   \\
    EEIL \cite{castro2018end}  & 68.68 &	53.63 &	47.91 &	44.20 &	36.30 &	27.46 &	25.93 &	24.7 &	23.95 &	24.13 &	22.11	 & 46.57 &	36.27   \\
    NCM \cite{hou2019learning}  & 68.68 &	57.12 &	44.21 &	28.78 &	26.71 &	25.66 &	24.62 &	21.52 &	20.12 &	20.06 &	19.87 &	48.81 &	32.49   \\
    D-Cosine \cite{vinyals2016matching}  & 75.90 &	70.21 &	65.36 &	60.14 &	58.79 &	55.88 &	53.21 &	51.27 &	49.38 &	47.11 &	45.67 &	30.23 &	57.54  \\
    TOPIC \cite{tao2020few} & 68.68 &	62.49 &	54.81 &	49.99 &	45.25 &	41.40 &	38.35 &	35.36 &	32.22 &	28.31 &	26.28 &	42.40 &	43.92   \\
    CEC$^{*}$ \cite{zhang2021few} & 75.68 &	71.82 &	68.44 &	63.53 &	62.48 &	58.20 &	57.51 &	55.61 &	54.72 &	53.34 &	52.12 &	23.56 &	61.22   \\
    MetaFSCIL \cite{chi2022metafscil} & 75.90 &	72.41 &	68.78 &	64.78 &	62.96 &	59.99 &	58.30 &	56.85 &	54.78 &	53.82 &	52.64 &	23.26 &	61.93   \\
    FACT \cite{zhou2022forward} & 75.90 &	73.23 &	70.84 &	66.13 &	65.56 &	62.15 &	61.74 &	59.83 &	58.41 &	57.89 &	56.94 &	18.96 &	64.42   \\
    ERDR \cite{liu2022few} & 75.90 &	72.14 &	68.64 &	63.76 &	62.58 &	59.11 &	57.82 &	55.89 &	54.92 &	53.58 &	52.39 &	23.51 &	61.52   \\
    DBONet \cite{guo2023decision} & 78.66 &	75.53 &	72.72 &	69.45 &	67.21 &	65.15 &	63.03 &	61.77 &	59.77 &	59.01 &	57.42 &	21.24 &	66.34    \\
    BiDist$^{*}$ \cite{zhao2023few} & 77.48 &	73.75 &	70.23 &	66.66 &	65.27 &	62.27 &	60.36 &	60.07 &	57.72 &	56.73 &	55.85 &	21.63 &	64.22   \\
    SAVC$^{*}$ \cite{song2023learning} & 81.57 &	77.64 &	74.80 &	70.38 &	69.96 &	67.13 &	66.46 &	65.26 &	63.52 &	62.92 &	62.36 &	19.21 &	69.27   \\
    NC-FSCIL$^{*}$ \cite{yang2023neural} & 79.85 &	73.56 &	70.20 &	66.52 &	63.28 &	60.59 &	59.34 &	57.83 &	55.49 &	54.11 &	52.80 &	27.05 &	63.05   \\

    \midrule
    ViT finetune  & 84.32 & 	74.88 & 	69.23 & 	60.96 & 	47.15 & 	42.63 & 	36.22 & 	32.18 & 	26.15 & 	23.71 & 	25.45 & 	58.87 & 	47.53 \\
    L2P$^*$ \cite{0002ZL0SRSPDP22} & 84.26 & 	43.99 & 	29.88 & 	22.64 & 	18.36 & 	14.79 & 	12.57 & 	11.55 & 	10.60 & 	9.37 & 	11.06 & 	73.20 & 	24.46  \\
    DualPrompt$^*$ \cite{0002ZESZLRSPDP22} & 84.53 & 	61.88 & 	52.56 & 	41.79 & 	35.65 & 	33.46  &	31.16 & 	25.80 & 	21.90 & 	18.10 & 	16.67 & 	67.86 & 	38.50 \\
   \midrule
   \rowcolor{blue!10} LGP-CAG2G(GAT)  & 84.61  & 81.67  & \textbf{82.45}  & \textbf{79.83}  & 76.41  & 75.64  & \textbf{77.72}  & \underline{75.46}  & 71.46  & 71.56  & 72.98 & 11.63  & 77.25 \\
   \rowcolor{blue!10}LGP-CAG2G(GraphSage) & \textbf{85.47} & 	\textbf{82.41} & 	82.19 & 	79.48 & 	\textbf{78.38} & 	\underline{76.63} & 	77.22 & 	75.23 & 	\underline{72.90} & 	\underline{72.48} & 	\underline{74.23} & 	11.24 & 	\underline{77.87} \\
   \rowcolor{blue!10} \textbf{LGP-CAG2G(GATv2)} & \underline{85.37} & 	\underline{82.18} & 	\underline{82.33} & 	\underline{79.59} & 	\underline{77.76} & 	\textbf{76.86} & 	\underline{77.37} & 	\textbf{76.93} & 	\textbf{73.49} & 	\textbf{72.82} & 	\textbf{74.87} & 	10.50  & 	\textbf{78.14} \\
   \rowcolor{blue!10}LGP-CAG2G(GCN) & 84.84 & 	81.36 & 	82.10 & 	77.66 & 	76.19 & 	75.68 & 	76.12 & 	72.86 & 	68.57 & 	68.54 & 	70.45 & 	14.39 & 	75.85 \\
   \rowcolor{blue!10}LGP-CAG2G(GCNII) & 79.64 & 	77.98 & 	77.37 & 	73.52 & 	73.26 & 	71.73 & 	71.22 & 	70.45 & 	71.27 & 	69.55 & 	69.52 & 	\underline{10.12} & 	73.23 \\
   \rowcolor{blue!10}LGP-CAG2G(PairNorm) & 81.84 & 	78.49 & 	80.76 & 	76.74 & 	75.79 & 	75.13 & 	75.10 & 	72.90 & 	69.67 & 	70.59 & 	71.80 & 	\textbf{10.04} & 	75.35 \\
   \rowcolor{blue!10}LGP-CAG2G(GGCN) & 80.69 & 	78.19 & 	79.68 & 	75.94 & 	74.81 & 	73.18 & 	72.17 & 	69.56 & 	68.12 & 	68.23 & 	69.33 & 	11.36 & 	73.63 \\
    \bottomrule
    \end{tabular}%
    }
    \caption{Comparison with SOTA methods on CUB200 dataset for FSCIL. $*$ are reproduced following the settings of the original paper. The remaining experimental results come from CEC\protect\cite{zhang2021few}, NC-FSCIL\protect\cite{yang2023neural} or the results reported in the original paper.}
  \label{tab:CUB200}%

\end{table*}%

\begin{table*}[!t]
  \centering
    \setlength{\tabcolsep}{2.8mm}
\renewcommand{\arraystretch}{1.0}
\resizebox{0.9\linewidth}{!}{
    \begin{tabular}{cccccccccccccc}
    \toprule
    \multirow{2}{*}{Method} & \multicolumn{11}{c}{Acc. in each session (\%) ↑}                                      & \multirow{2}{*}{PD $\downarrow$} & \multirow{2}{*}{\shortstack{Average \\ ACC ↑}}\\
\cmidrule{2-12}          & 0     & 1     & 2     & 3     & 4     & 5     & 6     & 7     & 8    & 9 & 10 &  & \\
    \midrule
    iCaRL$^*$ \cite{rebuffi2017icarl} & 55.74 &	38.49 &	36.67 &	33.90 &	24.79 &	20.33 &	17.83 &	16.73 &	14.09 &	13.53 &	10.80 &	44.95 &	25.72   \\
    EEIL$^*$ \cite{castro2018end}  & 55.74 &	39.47 &	35.97 &	33.94 &	24.47 &	18.27 &	15.93 &	15.17 &	14.03 &	13.77 &	11.75	 & 44.00 &	25.32   \\
    NCM$^*$ \cite{hou2019learning}  & 55.74 &	42.96 &	32.27 &	18.52 &	14.88 &	16.47 &	14.62 &	11.99 &	10.20 &	9.70 &	9.51 &	46.24 &	21.53   \\
    D-Cosine$^*$ \cite{vinyals2016matching}  & 62.96 &	56.05 &	53.42 &	49.88 &	46.96 &	46.69 &	43.21 &	41.74 &	39.46 & 36.75 &	35.31 &	27.66 &	46.59  \\
    TOPIC$^*$ \cite{tao2020few} & 55.74 &	48.33 &	42.87 &	39.73 &	33.42 &	32.21 &	28.35 &	25.83 &	22.30 &	17.95 &	15.92 &	39.83 &	32.97   \\
    CEC$^{*}$ \cite{zhang2021few} & 64.74 & 	57.66 & 	55.25 &  52.05 & 	49.35 & 	47.97 & 	46.38 & 	44.22 & 	42.42 & 	41.03 & 	39.59 & 	25.15 & 	49.15 \\
    MetaFSCIL$^*$ \cite{chi2022metafscil} & 65.14 &	56.66 &	53.86 &	49.82 &	45.23 &	43.10 &	40.29 &	37.14 &	35.11 &	33.11 &	30.73 &	34.41 &	44.56   \\
    FACT$^*$ \cite{zhou2022forward} & 65.14 &	57.48 &	55.92 &	51.17 &	47.83 &	45.26 &	43.73 &	40.12 &	38.74 &	37.18 &	35.03 &	30.11 &	47.05   \\
    ERDR$^*$ \cite{liu2022few} & 65.14 & 56.39 & 53.72 &	48.80 &	44.85 &	42.22 &	39.81 &	36.18 &	35.25 &	32.87 &	30.48 &	34.66 &	44.16   \\
    DBONet$^*$ \cite{guo2023decision} & 67.90 &	59.78 &	57.80 &	52.49 &	49.48 &	48.26 &	45.02 &	42.06 &	40.10 &	38.30 &	35.51 &	32.39 &	48.97    \\
    BiDist$^{*}$ \cite{zhao2023few} & 31.97 & 	28.15 & 	26.02 & 	23.77 & 	22.24 & 	21.09 & 	19.86 & 	18.59 & 	17.41 & 	16.28 & 	15.47 & 	\textbf{16.50} & 	21.90 \\
    SAVC$^{*}$ \cite{song2023learning} & 68.81 & 60.55 & 58.63 & 	54.20 & 	50.94 & 	49.20 & 	47.32 & 	44.43 & 	42.92 & 	41.15 & 	39.70 &	29.11 & 	50.71 \\ 
    NC-FSCIL$^{*}$ \cite{yang2023neural} & 59.48 & 	51.31 & 	48.67 & 	45.10 & 	41.70 & 	40.27 & 	38.34 & 	36.22 & 	34.86 & 	32.63 & 	31.42 & 	28.06 & 	41.82 \\ 

    \midrule
    ViT finetune  & 64.88 & 	54.02 & 	49.45 & 	44.03 & 	40.08 & 	35.56 & 	31.49 & 	28.88  & 	28.19 & 	26.13 & 	23.04 & 	41.83 & 	38.70 \\
    L2P$^*$ \cite{0002ZL0SRSPDP22}& 63.93 & 	31.45 & 	20.70 & 	14.67 & 	11.05 & 	8.97 & 	7.80 & 	6.31 & 	5.60 & 	4.60 & 	3.92 & 	60.01 & 	16.27 \\ 
    DualPrompt$^*$ \cite{0002ZESZLRSPDP22} & \underline{69.57} & 	40.13 & 	30.46 & 	21.51 & 	16.22 & 	14.74 & 	12.42 & 	12.85 & 	12.58 & 	10.29 & 	8.42 & 	62.15 & 	22.74 \\ 
    
    \midrule
    \rowcolor{blue!10}LGP-CAG2G  & \textbf{69.69}  & 62.10  & 59.29  & 55.88  & 51.90  & 49.28  & 47.75  & 47.05  & 45.41  & 44.09  & 41.95 & 27.74  & 52.21 \\
    \rowcolor{blue!10}LGP-CAG2G(GraphSage) & 67.61 & 	\underline{62.23} & 	\underline{59.83} & 	\underline{55.96} & 	\underline{52.72} & 	\underline{52.04} & 	\underline{49.78} & 	\underline{48.59} & 	\underline{47.88} & 	\underline{46.14} & 	\underline{44.73} & 	\underline{22.88} & 	\underline{53.41} \\
    \rowcolor{blue!10} \textbf{LGP-CAG2G(GATv2)} & 67.32 & 	\textbf{62.40} & 	\textbf{60.25} & 	\textbf{56.08} & 	\textbf{53.27} & 	\textbf{52.35} & 	\textbf{50.31} & 	\textbf{48.98} & 	\textbf{48.21} & 	\textbf{46.31} & 	\textbf{45.05} & 	22.92 & 	\textbf{53.74} \\
    \rowcolor{blue!10}LGP-CAG2G(GCN) & 66.55 & 	60.96 & 	57.64 & 	52.59 & 	50.11 & 	48.42 & 	46.52 & 	45.38 & 	44.15 & 	41.61 & 	40.02 & 	26.53 & 	50.36 \\
   \rowcolor{blue!10}LGP-CAG2G(GCNII) & 65.33 & 	59.64 & 	55.92 & 	51.71 & 	48.41 & 	47.50 & 	44.05 & 	42.49 & 	41.44 & 	38.95 & 	37.03 & 	28.30 & 	48.41 \\
   \rowcolor{blue!10}LGP-CAG2G(PairNorm) & 66.45 & 	59.76 & 	56.77 & 	52.34 & 	49.12 & 	47.70 & 	45.31 & 	43.69 & 	42.93 & 	40.70 & 	38.78 & 	27.67 & 	49.41 \\
   \rowcolor{blue!10}LGP-CAG2G(GGCN) & 65.99 & 	59.80 & 	56.45 & 	52.13 & 	48.87 & 	47.70 & 	44.78 & 	43.19 & 	42.29 & 	39.93 & 	38.01 & 	27.98 & 	49.01 \\
    
    \bottomrule
    \end{tabular}%
    }
    \caption{FSCIL comparison on ImageNet-R dataset. Results of methods with $*$ are reproduced following the settings of the original paper.}
  \label{tab:imagenet_r}%

\end{table*}%

\subsection{Experimental Setup}
\paragraph{Backbone}Following \cite{0002ZL0SRSPDP22,0002ZESZLRSPDP22}, we choose the pretrained ViT \cite{DosovitskiyB0WZ21} trained on Imagenet-1K as the frozen feature extractor $f_{\boldsymbol \varphi}$. The rest of the methods' backbone settings were referenced from their original papers.

\paragraph{Datasets}Following the settings in \cite{zhang2021few}, we conduct tests on two widely-used datasets in FSCIL: CIFAR100 \cite{krizhevsky2009learning}, CUB200 \cite{wah2011caltech}. Furthermore, to avoid overlap between the pretraining dataset of the feature extractor and the dataset used for evaluation, ensuring a fair comparison, we do not use the commonly used miniImageNet dataset for evaluation. Instead, we utilize the more challenging ImageNet-R dataset \cite{HendrycksBMKWDD21}, as done in \cite{0002ZESZLRSPDP22}.

\textbf{CIFAR100.} The CIFAR100 dataset comprises $100$ classes, each with $500$ training images and $100$ test images, all of size $32\times 32$. Out of these, $60$ classes serve as base classes, the remaining are split into 8 novel incremental sessions with only $5$ training samples per class, i.e. $8$-way $5$-shot.

\textbf{CUB200.} The CUB200 dataset is a fine-grained image classification dataset containing $200$ classes of birds, comprising around $6000$ training images and $6000$ testing images with size $224\times 224$. Out of the whole $200$ classes, $100$ of them are designated as base classes, while the remaining $100$ classes are distributed across $10$ new sessions, each with $5$ training images for every class, i.e. $10$-way $5$-shot.

\textbf{ImageNet-R.} The ImageNet-R dataset consists of $200$ classes with $24000$ training images and $6,000$ test images. The images from this dataset include different styles such as cartoon, graffiti, and origami, as well as challenging samples from ImageNet that standard training models struggle to classify. It exhibits significant intra-class diversity, making it difficult for rehearsal-based incremental learning methods to effectively work with a small number of rehearsal images \cite{0002ZESZLRSPDP22}. Out of the $200$ classes, $100$ are used to construct the base session, while the remaining $100$ classes are divided into $10$ incremental sessions. Each new class contains $5$ training samples, i.e. $10$-way $5$-shot.

\paragraph{Baselines}To ensure a comprehensive evaluation, we select several classic CIL methods, including iCaRL \cite{rebuffi2017icarl}, EEIL \cite{castro2018end}, NCM \cite{hou2019learning}, and D-Cosine \cite{vinyals2016matching}, as well as competitive FSCIL methods TOPIC \cite{tao2020few}, CEC \cite{zhang2021few}, MetaFSCIL \cite{chi2022metafscil}, FACT \cite{zhou2022forward}, ERDR \cite{liu2022few}, and several SOTA FSCIL methods DBONet \cite{guo2023decision}, BiDist \cite{zhao2023few}, SAVC \cite{song2023learning}, and NC-FSCIL \cite{yang2023neural}. Additionally, we evaluate the ViT model in a fine-tuned manner on each task with frozen feature extraction and same rehearsal buffer size as our method, to confirm the gains brought by our method. Considering that no FSCIL method with ViT as the backbone has been proposed, we select two CIL SOTA methods, L2P \cite{0002ZL0SRSPDP22} and DualPrompt \cite{0002ZESZLRSPDP22}, based on pre-trained ViT for comparison to demonstrate the advantage of our method in FSCIL with the same backbone.

\paragraph{Implementation Details} 
When evaluating our method, all samples were resized to $224\times 224$ to accommodate ViT's input requirements. The dimension of the interactive feature $d_\xi$ is set to $1024$, and the number of segments for local features $S$ is set to $8$. The hyperparameters $\lambda$ and $\eta$ are both set to $0.1$. We only store one sample per class for rehearsal purposes to verify the effectiveness of our method under extremely limited buffer conditions. The Adam optimizer is chosen with a learning rate of $1e-4$, to train the model on the base session and each incremental session for $100$ epochs, following the setup described in \cite{zhang2021few}. 

For the experiments on the ImageNet-R dataset, we reproduced all the compared methods. For the experiments on CIFAR100 and CUB200 datasets, we reproduced the results of SOTA methods including CEC, BiDist, SAVC, NC-FSCIL, as well as L2P and DualPrompt, and reported the reproduced results. The results of other methods were taken from their original papers. All the reproduced methods were implemented with the original recommended settings from their respective papers.

\begin{figure*}[ht]
\centering 
\includegraphics[width=0.9\textwidth]{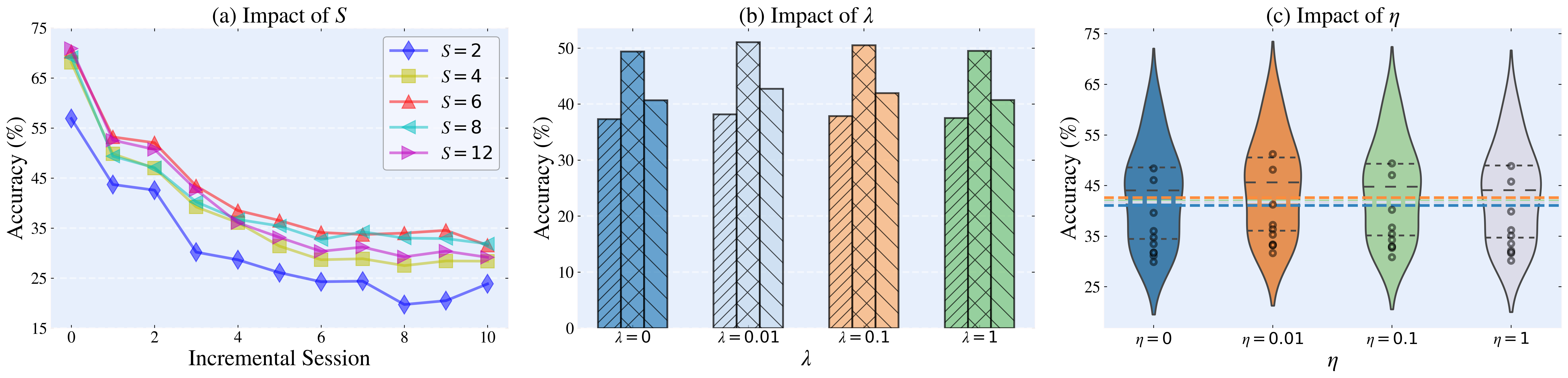}

\caption{The results of ablation experiments on ImageNet-R are shown. (a) demonstrates the performance of our method on the evaluation set $\mathcal E^{(t)}$ of all previous sessions under different choices of $S$. The three bars in each group of (b) display, from left to right, the average accuracy of our method on each incremental sessions for the corresponding choice of $\lambda$, the average accuracy on the evaluation set $\mathcal E^{(t)}$ after each round of incremental session, and the final evaluation set accuracy after the last session. We draw a violin plot of the test precision for each round of few-shot incremental sessions for each $\eta$ choice in (c), where the dashed line indicates the final test precision on $\mathcal E^{(T)}$ after completing all incremental training.}
\label{Fig3}
\end{figure*}

\begin{figure*}[t]
\centering 
\includegraphics[width=0.9\textwidth]{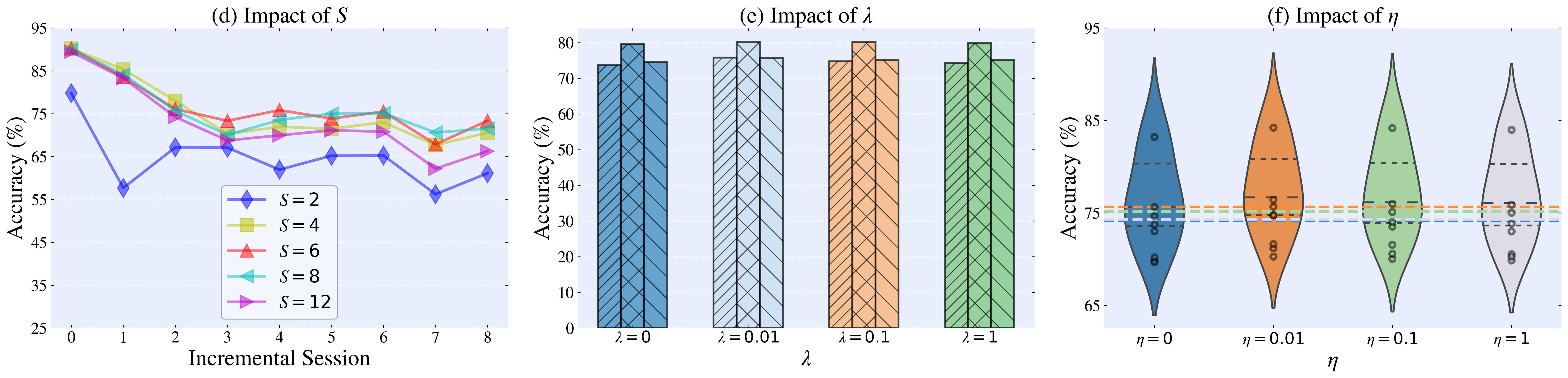}
\includegraphics[width=0.9\textwidth]{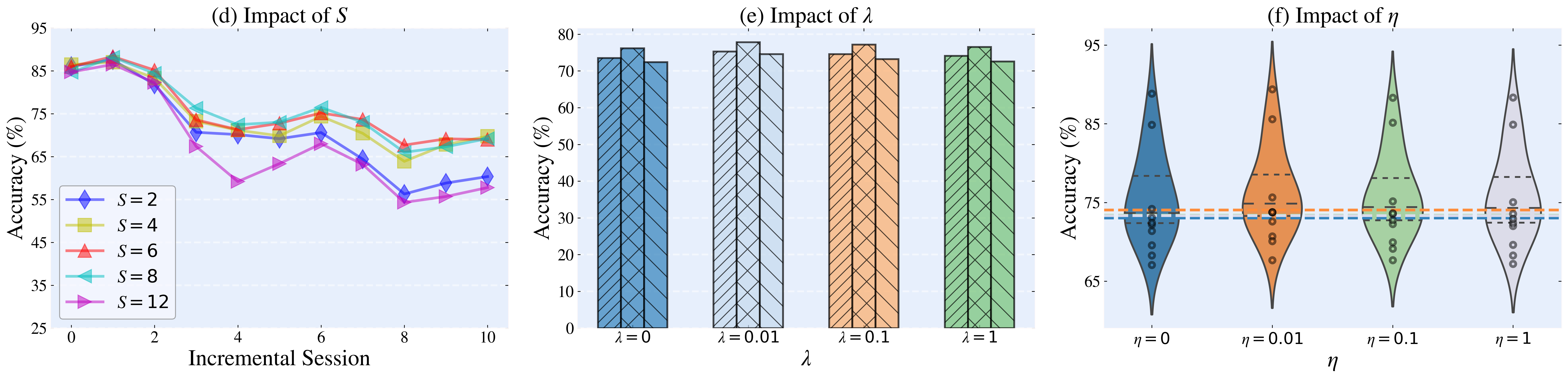}
\caption{The results of ablation experiments on CIFAR100 (three figures above) and CUB200 (three figures below) datasets. (a) and (d) demonstrates the performance of our method on the evaluation set $\mathcal E^{(t)}$ of all previous sessions under different choices of $S$. The three bars in each group of (b) and (e) display, from left to right, the average accuracy of our method on each incremental sessions for the corresponding choice of $\lambda$, the average accuracy on the evaluation set $\mathcal E^{(t)}$ after each round of incremental session, and the final evaluation set accuracy after the last session. We draw a violin plot of the test precision for each round of few-shot incremental sessions for each $\eta$ choice in (c) and (f), where the dashed line indicates the final test precision on $\mathcal E^{(T)}$ after completing all incremental training.}
\label{Fig4}
\end{figure*}

\noindent \textbf{Evaluation Protocol.} 
Following the settings in \cite{zhang2021few}, after each session, we assess the model by testing it on the evaluation set $\mathcal{E}^{(t)}=\cup_{s=1}^{t} \mathcal W^{(s)}$, which consisted of all the previous session test sets after each session, and reporting the Top $1$ accuracy. Additionally, we define a performance dropping rate (PD) to quantify the absolute accuracy degradation in the last session, given by $ \text{PD}=\mathcal{C}_0-\mathcal{C}_N $, where $\mathcal{C}_0$ represents the classification accuracy in the base session, and $\mathcal{C}_N$ is the accuracy in the last session.

\subsection{Comparison with State of The Art Methods}
\label{Comparison}

We conduct comparative experiments of all methods on CIFAR100, CUB200, and ImageNet-R datasets, results are shown in Tab. \ref{tab:CIFAR100}, Tab. \ref{tab:CUB200}, and Tab. \ref{tab:imagenet_r} respectively. Our method outperforms other methods in terms of average accuracy and final accuracy on all sessions. Specifically, even in the case of having only one image per class stored in the rehearsal buffer, our method achieves an average accuracy improvement of $13.71\%$ and $7.97\%$, as well as a final accuracy improvement of $19.04\%$ and $10.62\%$ compared to the second-best method on CIFAR100 and CUB200 datasets, while demonstrating the lowest performance dropping rate. 

\begin{figure*}[htbp]

\centering 
\includegraphics[width=0.95\textwidth]{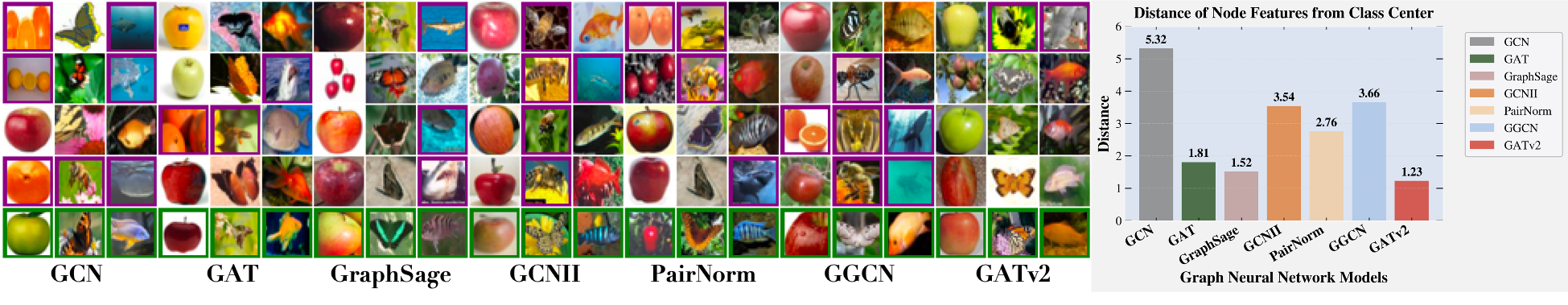}
\caption{
The histogram illustrates the distances between the node features learned by different GNN models and their class centers. The right figure displays the distances between the anchor points learned by different GNN models and the actual samples, where the samples marked in green are the actual anchor samples and the samples marked in purple do not belong to this category.
}
\label{Fig5}

\end{figure*}

On the ImageNet-R dataset, our method shows a $1.27\%$ and $2.25\%$ improvement in average accuracy and final accuracy respectively compared to the second-best method. In terms of performance dropping, although our method does not achieve the best, BiDist and CEC, which have the best and second-best performance dropping rate, respectively, both have significantly lower test accuracy in each round of session compared to our method. Additionally, it is worth noting that our method shows significant improvements in both average accuracy and final accuracy compared to ViT Finetune with the same rehearsal buffer size, which further validates the practical gains brought by our strategy. 
In addition, we randomly tested several different graph neural networks to assess the impact of capturing geometric structure information on the output characteristics of FSCIL. Through experiments, we found that utilizing graph neural networks for local feature interactions can significantly enhance the model's performance in FSCIL. Furthermore, we observed that the model performs best when using GAT and other spatial-based graph neural networks, including GraphSage, GATv2 \cite{Velickovic2017GraphAN, hamilton2017inductive, brody2021attentive, wu2020comprehensive}. This may be because spatial-based models execute local graph convolutions on each node, allowing easy weight sharing between different positions and structures. Compared to spectrum-based GNN models such as GCN, PairNorm, GCNII, GGCN \cite{kipf2016semi, zhao2019pairnorm, chen2020simple, yan2022two, wu2020comprehensive}, spatial-based GNN models can better capture the local structure of nodes and relationships between adjacent nodes, thus learning superior node representations.
To validate this perspective, we conducted further experiments. Due to the relatively small scale of the graphs used in the FSCIL problem setting, we randomly generated a graph consisting of only 6 nodes for experimentation. We compared the distances of the node features learned by the selected 7 GNN models to the distances from the node to its class center as a measure of the model's representation ability for nodes. The experimental results are shown in Fig. \ref{Fig5}. We found that, compared to spectrum-based GNN models, spatial-based GNN models learned node features that are closer to the class center, indicating that spatial-based GNN models possess better node representation capabilities.
\subsection{Ablation Study}
\label{Ablation}
We conduct a confirmatory evaluation of the proposed components' configurations. 
\paragraph{Ablation Results for $S$ in G2G Interaction}We first perform FSCIL tasks on various datasets with different values of $S$ and recorded the accuracy on the cumulative test set $\mathcal E$ in each session to evaluate the impact of different granularity of local feature partitioning on our G2G interaction mechanism, as shown in Fig. \ref{Fig3} (a). It is worth noting that the performance is weakest when $S=2$, which is due to the fact that the graph constructed with only two nodes in $S=2$ contains insufficient graph information for local feature graph construction, thus limiting the learning of GAT. As $S$ increases, the model's performance improves significantly, achieving the best performance at $S=6$ and $S=8$. The catastrophic forgetting problem is greatly mitigated, confirming the effectiveness of our method. However, when $S=12$, the performance shows a slight decline compared to $S=6$ and $S=8$. This is because excessively fine-grained partitioning of global features leads to low-dimensional local features, making it difficult to generate class prototypes with sufficient inter-class margins in such a low-dimensional space to perform well in classification.
\paragraph{Ablation Results of $\lambda$ for LGP}
To verify the improvement of LGP mechanism on the model's few-shot generalization and memory capacity, we recorded the performance of the model on the test set $\mathcal W$ of each session under different $\lambda$ choices, as well as the accuracy of the model on the final evaluation set $\mathcal E$ after training is completed, as shown in Fig. \ref{Fig3} (b). When $\lambda=0.01$, LGP can bring an average improvement of about $1\%$ to the model's testing accuracy on incremental few-shot classes. At the same time, because it further constrains the geometric structure of local features, it introduces clearer graph structural information for G2G interaction, making it possible to more accurately restore features of previous classes during rehearsal and alleviate catastrophic forgetting.
\paragraph{Ablation Results of $\eta$ in Contrastive Augmentation}To evaluate the improvement of few-shot generalization ability brought by Contrastive Augmentation (CA) to our G2G strategy, we recorded the performance on each session's test set $W^{(t)}$ in each session $t$. To further demonstrate that the addition of CA does not limit the model's memory capacity, we recorded the accuracy of the model on the final evaluation set $\mathcal E$ after training is completed, as shown in Fig. \ref{Fig3} (c). The addition of CA effectively improved the model's accuracy on incremental classes with fewer samples, improving the model's few-shot generalization, especially at $\eta=0.01$ and $0.1$, while further alleviating catastrophic forgetting. However, when lambda is too large ($\eta=1$), the improvement of the model is not significant. This is because excessively large contrastive loss will make the model overly focus on augmented data, thereby affecting the model's training on normal samples.

\section{Additional Ablation Studies}

The results of the ablation experiments mentioned in Section \ref{Ablation} on the CIFAR100 and CUB200 datasets are presented in Figure \ref{Fig4}. To control variables, in experiments with variable $S$ (corresponding to (a) and (d) in Fig. \ref{Fig4}), we maintained $\lambda=0.1$ and $\eta=0.1$. In experiments with variable $\lambda$ (corresponding to (b) and (e) in Fig. \ref{Fig4}), we kept $S=8$ and $\eta=0.1$. In experiments with variable $\eta$ (corresponding to (c) and (f) in Fig. \ref{Fig4}), we maintained $S=8$ and $\lambda=0.1$. The ablation results on CIFAR100 and CUB200 exhibit a similar trend to the ablation results on ImageNet-R in Section \ref{Ablation}.

For G2G, the improvement in model performance is not significant when $S$ is too small or too large, while the model shows optimal performance when $S$ is set to $6$ or $8$. LGP achieves the best improvement when $\lambda=0.01$. Specifically, it provides an average improvement of $1.59\%$ (on CIFAR100) and $1.76\%$ (on CUB200) for each round of few-shot incremental sessions, an average improvement of $0.42\%$ (on CIFAR100) and $1.66\%$ (on CUB200) in accuracy on each round's evaluation set $\mathcal E^{(t)}$, and an improvement of $1.57\%$ (on CIFAR100) and $2.16\%$ (on CUB200) in accuracy on the final round's evaluation set $\mathcal E^{(T)}$. Regarding the comparison of augmentation strategies, the best performance is achieved when $\eta=0.01$. Specifically, it provides an average accuracy improvement of $1.11\%$ (on CIFAR100) and $1.10\%$ (on CUB200) for each few-shot incremental session and an improvement of $1.50\%$ (on CIFAR100) and $1.01\%$ (on CUB200) in accuracy on the final round's evaluation set $\mathcal E^{(T)}$.

\section{Conclusion}
 In this paper, we delved into the explicit memory's local interaction modeling in FSCIL. We introduced the efficient G2G interaction mechanism, extending the feature-prototype interaction to encompass local geometric structure. Additionally, the LGP mechanism was introduced to constrain thus strengthen the local geometric structure, and prevent feature collapse. Furthermore, we proposed the CAG2G interaction to enhance few-shot generalization by addressing sample scarcity in new sessions. Experimental comparisons across CIFAR-100, CUB200, and ImageNet-R datasets showcased the superiority of our method over competitive baselines.

\bibliographystyle{IEEEtran}
\bibliography{TCSVT}

\end{document}